# Plantbot: A New ROS-based Robot Platform for Fast Building and Developing

Liulong Ma and Yanjie Liu *

*State Key Laboratory of Robotics and System*
*Harbin Institute of Technology*
*Harbin, Heilongjiang, 15000, China*
marooncn@163.com and yjliu@hit.edu.cn

Yibo Gao and ChuanMing Zhu

*School of Mechanical Engineering*
*Tongji University*
*4800, Cao'an Highway, Shanghai 201804, China*
526348668@qq.com and 01065@tongji.edu.cn

*Abstract* – In recent years, the Robot Operating System (ROS) is developing rapidly and has been widely used in robotics research because of its flexible, open source, and easy-to-expand advantages. In scientific research, the corresponding hardware platform is indispensable for the experiment. In the field of mobile robots, PR2, Turtlebot2, and Fetch are commonly used as research platforms. Although these platforms are fully functional and widely used, they are expensive and bulky. What's more, these robots are not easily redesigned and expanded according to requirements. To overcome these limitations, we propose Plantbot, an easy-building, low-cost robot platform that can be redesigned and expanded according to requirements. It can be applied to not only fast algorithm verification, but also simple factory inspection and ROS teaching. This article describes this robot platform from several aspects such as hardware design, kinematics, and control methods. At last two experiments, SLAM and Navigation, on the robot platform are performed. The source code of this platform is available[1].

## I. INTRODUCTION

The Robot Operating System (ROS) has emerged as one of the most popular frameworks in robotics research and Robot Startups. On the one hand, ROS is used by startups to develop products rapidly and is gradually compatible with traditional industrial robots [1]. On the other hand, it is used to research cutting-edge robotics technologies such as:

1. simultaneous localization and mapping(SLAM): Gmapping [2], Hector SLAM [3], Cartographer[4], RGB-D SLAM [5];

2. autopilot: apollo, OSSDC;

3. reinforcement learning: rl-texplore-ros-pkg, gym-gazebo [6].

What's more, In December 2017, Open Source Robotics Foundation (OSRF), the maintainer of ROS, released ROS2 to fundamentally solve the problems of stability and security of ROS [7]. And the application of ROS will be more extensive. In robot experiment, researchers need to have a robot platform to do the test. PR2, Turtlebot2, and Fetch are the most widely used robot platforms in robotic research. These robots are developed initially to run ROS. So they are well supported to ROS. But the price of these robots is very high, specifically, PR2 priced at $340k, turtlebot2 priced at $2k. The functions of PR2 and Fetch are powerful yet for researchers and teachers, they may need only a flexible mobile platform with LiDAR to implement SLAM algorithm for tests or demonstrations. These robots are too expensive and bulky for researchers or robot teachers. And as a commercial product, it's not easy to redesign its structure or add new sensors to do the new experiment. For example, it is hard to change Kinect with LiDAR in PR2. So how to make a trade-off between price, flexibility, and function is an issue to be addressed.

We propose a new ROS-based robot platform, Plantbot, which you can build and develop easily by yourself.

## II. HARDWARE DESIGN

The hardware design of this robot platform is completed in Solidworks. The designed parts were processed by wire cutting and 3D printing. The design of the circuit includes the selection of the micro-controller, the power supply of each sensor, and the voltage conversion.

### A. The Mechanical Structure Design

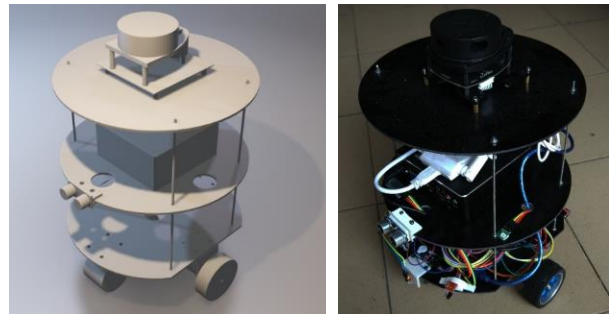

Fig. 1 Assembly drawing(left) and Physical Object(right)

Plantbot refers to the Turtlebot2 design using a three-layer structure. The lower board is used to place the circuit board and battery as a power supply and circuit system. The middle board places the core controller as a computing and control system. The upper board is used to place core sensors or operating equipment, such as LiDAR (Light Detection And Ranging), Kinect, robotic arm, as a sensing and manipulating system.

The circuit board connects the battery and various sensors for voltage conversion and sensor power supply. In Plantbot's design, ultrasonic sensors are used for obstacle avoidance, infrared sensors are used to prevent fall, DC motors are employed for driving, encoders are used for speed measurement, Bluetooth receiver module is served as the function to receive control commands, and LED lights are used to display the status of robots. These sensors are necessary to a robot platform. The more detailed circuit design will be described later.

---
[1] https://github.com/marooncn/plantbot

The selection of the core controller should be considered from several aspects. For example, the size of the controller is constrained to the design dimension of the middle layer board, and the performance of the controller is determined by the calculation amount and the real-time requirement. We choose Next Unit of Computing (NUC) [8] released by Intel as core controller. Because on the one hand NUC volume is smaller, less than 0.7 cubic decimetres; On the other hand, it has good computing performance, in our case, the i5-7260U processor is used, the computing performance is equivalent to a usual computer, which is enough for ordinary robot tasks.

The upper board placed core sensors in the robot mission. In the study of the laser SLAM, LiDAR is used, and Kinect is chosen in the visual SLAM study. Also, the Plantbot can be used for inspection or transportation at the factory. In this case, the upper board can be used to carry mechanical parts, which can relieve the burden of workers.

The entire three-dimensional solid design is completed in Solidworks. In the design process, the board is completed first. The board size is decided by the task, and holes on the board are reserved for installing sensors or allowing wires to pass through. After the various components are designed, they can be assembled in the software. After confirming that they are correct, parts that need to be processed can be exported. This entire design process ensures that this robot platform can be redesigned easily according to specific requirements.

*B. Parts Processing and Assembly*

The hardware design of this robot platform uses a wide range of machining and manufacturing technologies, such as wire cutting and 3D printing. The corresponding devices can be found easily in robotic labs or machining centres, or can be customized and processed at a very low price. For novices in mechanical design and manufacturing, the use of the related equipment is also very easy to start and relatively safe to operate, unlike the cutting machine tools, which are costly, difficult to operate, and prone to accidents for beginners.

Glass fiber is a low-cost and high-strength material, it is widely used in quadrotors and mobile robots. Glass fiber board is suitable as support boards in robot platform. Thus the glass fiber board is selected as raw material and is processed by wire cutting technology. Wire cutting is very suitable for processing plate-like parts, with the advantages of high speed and high precision.

The parts that connect sensors can be processed using 3D printing technology after they are designed. 3D printing refers to processes in which material is joined or solidified under computer control to create a three-dimensional object, with material being added together (such as liquid molecules or powder grains being fused together). It is very convenient to use 3D printer to process parts, and in most mechanical centres, it is easy to find a 3D printer. In Plantbot, the connectors used to connect the ultrasonic sensor and infrared sensors are processed by 3D printing technology, as shown in Fig. 2.

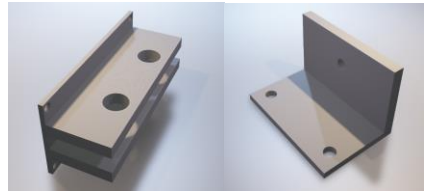

Fig. 2 3D printing parts

After the designed parts are processed, they can be assembled. The boards are connected by double studs and then fixed by nuts, one can easily change the position of the nut to change the distance between two boards. Next, all sensors, motors, wheels, and other purchased parts are installed, and the mechanical part of the robot is completed.

*C. Circuit Design*

The battery supplies all the components of the platform. Wiring from the battery is divided into two branches. One is connected to NUC. NUC requires 12V power supply, so a 12V 10Ah battery is selected; the other connects with the DC transformer, the transformer can convert 12V DC power to 5V DC power as the power of Micro Control Unit(MCU).

Table 1 Hardware of Plantbot

| Hardware | Function |
| --- | --- |
| NUC | Core Controller running Ubuntu 14.04 |
| Launchpad | MCU, connecting motor drives and encoders |
| Arduino | MCU, connecting Bluetooth receiver, infrared sensors and LED indicator |
| Motor driver | L298N (H bridge) |
| DC Motor | ASLONG JGB37, rated voltage is 12v |
| LiDAR | RPLIDAR A1, produced by SLAMTEC |

Plantbot uses two different MCUs, as shown in Table 1. Launchpad is responsible for receiving the PWM wave and outputting the corresponding voltage to DC motor driver to control the motor's steering and rotation speed. Launchpad gets the distance information of the ultrasonic sensor and sends it to NUC through the serial port after processing. Launchpad also receives the control commands published from NUC through the serial port and sends the command to the driver board.

Arduino receives data from infrared sensors and the Bluetooth communication module and sends processed information to NUC through the serial port. At the same time, it receives feedback of NUC and controls the LED indicator indicating the state of Plantbot. NUC provides a 12v USB interface that can be used to power the LiDAR.

III. KINEMATICS

The differential drive system is commonly used in mobile robot platform. Its mechanical structure is simple and it is often used in indoor navigation. The differential drive system consists of two coaxial drive wheels and one or two caster wheels for support.

*A. Mobile Robot Motion Description*

The mobile robot using ROS specifies that the coordinate axes of its local coordinate system are determined by the right-hand rule, that is, the x-axis points in the robot's forward direction and the y-axis points to the left side of the robot, and the z-axis upward. ROS publishes motion commands using the Twist message type. Topic name of the planar speed command is usually */cmd_vel*. After the node of the low-level controller subscribes to this topic, it can receive the messages of this topic, and control the platform to do the corresponding movement. In order to describe the motion of the robot, in addition to the local coordinate system, the global coordinate system also needs defining to determine its absolute position.

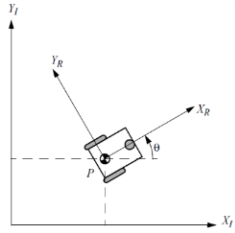

Fig. 3 Global and local coordinate system

As shown in Fig. 3, $X_I$ and $Y_I$ define the global coordinate system $\{X_I, Y_I\}$. $X_R$ and $Y_R$ define the local coordinate $\{X_R, Y_R\}$ with the center of mass $P$ of the mobile robot as the origin. Assume coordinate of the point $P$ in the global coordinate is (x, y) and the rotation angle of the local coordinate relative to the global coordinate system is $\theta$. We can use three elements to describe the pose of the robot in the global coordinate system. As follows

$$\xi_I = \begin{bmatrix} x \\ y \\ \theta \end{bmatrix} \quad (1)$$

To describe the movement of the platform, we need to map the movement in the global coordinate to the local. This map is a function describing the current pose and it can be determined using a rotation matrix [9]:

$$R(\theta) = \begin{bmatrix} \cos\theta & \sin\theta & 0 \\ -\sin\theta & \cos\theta & 0 \\ 0 & 0 & 1 \end{bmatrix} \quad (2)$$

The speed command of the global coordinate system can be converted into the local coordinate system.

$$\xi_R' = R(\theta)\xi_I' = \begin{bmatrix} \cos\theta & \sin\theta & 0 \\ -\sin\theta & \cos\theta & 0 \\ 0 & 0 & 1 \end{bmatrix} \begin{bmatrix} x' \\ y' \\ \theta' \end{bmatrix} = \begin{bmatrix} x_R' \\ y_R' \\ \theta_R' \end{bmatrix} \quad (3)$$

In this way, the speed command of the global coordinate can be mapped directly into the local coordinate and be sent to the platform to respond accordingly. Such a conversion process occurs during the navigation.

*B. Synthesis and Decomposition of Speed*

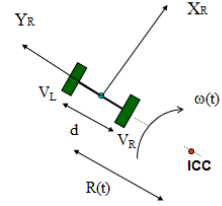

Fig. 4 Motion analysis (only two driving wheels are drawn)

In the movement of the mobile robot, the motion command of the platform received is the speed of the whole rigid body. In the differential drive system, the movement is driven by two driving wheels. Therefore, it is necessary to decompose the speed command of the platform into the speed of two driving wheels, which is the decomposition of the speed. The two driving wheel speeds calculated by encoders need to be combined into the speed of the whole rigid body to obtain the current movement speed, which is the synthesis of the speed. Let the synthesis speed of the robot be $(x_R', y_R', \theta_R')$, which is the linear velocity along the x and y axes and the angular velocity along the z axis in the local coordinate system. The left and the right wheel speeds are $V_L$ and $V_R$ respectively. The analysis of any small movement is as Fig. 4. In this figure, ICC is the instantaneous centre of curvature, R is the radius of rotation, w is the rotational angular velocity, and d is the centre distance of the two driving wheels. Analysis of motions relations [10]:

$$\begin{aligned} w(R + d/2) &= V_L \\ w(R - d/2) &= V_R \\ wR &= x_R' \end{aligned} \quad (4)$$

From the above formulas, we can solve the rotation radius $R$ and the rotation angular velocity $w$:

$$\begin{aligned} w &= (V_R - V_L)/d \\ R &= d(V_R + V_L)/(V_R - V_L) \end{aligned} \quad (5)$$

Substituting (5) into (4) to solve $x_R'$, the speed of the platform along x axis:

$$x_R' = (V_R + V_L)/2 \quad (6)$$

Plantbot is a non-holonomic restraint system, so its speed along the y-axis is zero, and the angular speed along the z-axis is equal to $w$, ie:

$$\begin{aligned} y_R' &= 0 \\ \theta_R' &= w = (V_R - V_L)/d \end{aligned} \quad (7)$$

Therefore, the speed of the synthesis can be calculated by the formula (6) and (7), and the decomposition of the speed from these two equations are:

$$\begin{aligned} V_L &= x_R' - \theta_R' \times d/2 \\ V_R &= x_R' + \theta_R' \times d/2 \end{aligned} \quad (8)$$

So the formulas for the synthesis and decomposition of speed are obtained.

## C. Positive Kinematics of Mobile Robot

During the localization of the mobile robot, it is necessary to know its absolute position in the global coordinate. $V_R$ and $V_L$ can be calculated using encoders. According to formula (3), the mapping relation between global coordinate and local coordinate, the robot's speed in the global coordinate is:

$$\xi_I' = R(\theta)^{-1}\xi_R' = \begin{bmatrix} \cos\theta & -\sin\theta & 0 \\ \sin\theta & \cos\theta & 0 \\ 0 & 0 & 1 \end{bmatrix} \begin{bmatrix} x_R' \\ y_R' \\ \theta_R' \end{bmatrix} \quad (9)$$

Substituting (6), (7) into (9):

$$\xi_I' = \begin{bmatrix} \cos\theta & -\sin\theta & 0 \\ \sin\theta & \cos\theta & 0 \\ 0 & 0 & 1 \end{bmatrix} \begin{bmatrix} (V_R+V_L)/2 \\ 0 \\ (V_R-V_L)/d \end{bmatrix} \quad (10)$$

Let the robot pose in the global coordinate system at the initial time:

$$\xi_{Io} = \begin{bmatrix} x_o \\ y_o \\ \theta_o \end{bmatrix} \quad (11)$$

Integrate formula (10) with the initial condition (11) to get the position in the global coordinate system:

$$\xi_I = \begin{bmatrix} x \\ y \\ \theta \end{bmatrix} = \begin{bmatrix} \int_t [\cos\theta_t (V_{Rt}+V_{Lt})/2]dt + x_o \\ \int_t [\sin\theta_t (V_{Rt}+V_{Lt})/2]dt + y_o \\ \int_t [(V_{Rt}-V_{Lt})/d]dt + \theta_o \end{bmatrix} \quad (12)$$

The variable with subscript t changes over time and represents the value of the variable at time t. For a physical system, the collected data are all discrete values, so the following formula is usually used:

$$\xi_I = \begin{bmatrix} x \\ y \\ \theta \end{bmatrix} = \begin{bmatrix} \sum_t [\cos\theta_t (V_{Rt}+V_{Lt})/2]\Delta t + x_o \\ \sum_t [\sin\theta_t (V_{Rt}+V_{Lt})/2]\Delta t + y_o \\ \sum_t [(V_{Rt}-V_{Lt})/d]\Delta t + \theta_o \end{bmatrix} \quad (13)$$

Where $\Delta t$ is the sampling time. If the value is too large, the calculation accuracy will be reduced. If it is set too small, it will make the calculation too much, beyond the core controller's capabilities, and will affect the result. Therefore, it is necessary to think about both the dynamic performance of the robot and the computing performance of the processor to choose an optical value.

As time goes, the positional error resulting from the above equation will increase. To avoid excessive accumulation of errors and improve positioning accuracy, when the mobile robot moves for a period, it is necessary to correct the position of the odometer. A common method is to set a marker or use other sensors for information fusion to correct the position.

## IV. LOW-LEVEL CONTROL

Low-level control is the implementation of the motion for Plantbot, which can be called directly in ROS after encapsulated.

### A. Low-level Node Graph

Low-level control of the robot platform includes command control and the control approaches to accurate execution. The low-level control is crucial to the robot and is a prerequisite for the robot to be able to perceive the upper layer accurately. Plantbot encapsulates the low-level control interface, which can be easily called by researchers. And the program is open source that researchers can change the low-level control code to achieve more precise control. Fig. 5 shows the whole node graph.

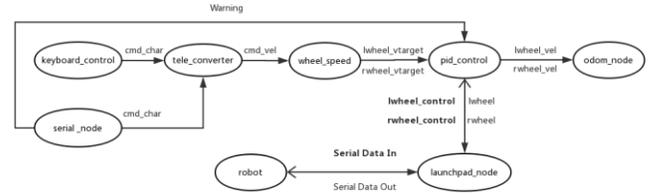

Fig. 5 Low-level node graph

keyboard_control and serial_control are two approaches initially provided by Plantbot to control the robot directly. The specific role of each node is as follows:

1. keyboard_control: This node represents that you can control the robot with specific keyboard keys, for example, 'w' represents straight forward motion.

2. serial_control: The interface of Arduino, it receives command from andriod App. The App is designed in MIT App Inventor [11]. This node can also receive the information of infrared sensors to achieve emergency stop to prevent falling.

3. tele_converter: Command conversion node, convert character commands to speed commands

4. wheel_speed: Speed decomposition node, decompose speed according to speed decomposition formula (8).

5. launchpad_node: The interface of Launchpad, it receives speed commands from speed controller to the motor driver, and simultaneously publishes the encoder information.

6. sec_control/pid_control: The controller of motor's speed. It will be explained in detail below.

7. odom_node: Odometer information measured by the encoders, calculated by the formula (13)

### B. Speed Controller

Plantbot provides two different controllers. pid_control uses the PID closed control and sec_control is an approach to segmented control.

PID is a simple and effective closed-loop control method that is widely used in industrial control. It is necessary to think about the dynamic response of the reactor and the specific requirements for use. Here we simulate PID control performance in Simulink based on motor parameters.

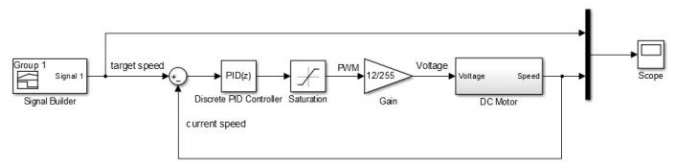

Fig. 6 Simulink Model

The PID controller is discrete and set the sampling time to 0.1 s. Signal Builder simulates the change of target speed. Considering the PWM output is 8 bits, set the upper limit of Saturation to 255 and the lower limit to -255. Turn on PID tuner [12] and adjust the response parameters to get the most suitable controller parameters. Use these parameters to get the best control response of PID controller as shown in Fig. 7.

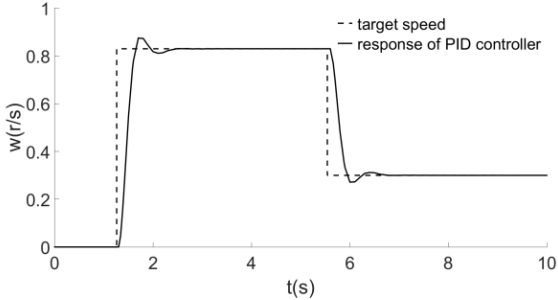

Fig. 7 The response of PID Controller

The rise time of the control response is within one second and the overshoot is less than 10% in our case. For a low-speed platform, when the target speed is updated at a relatively low frequency, the control effect can basically meet the requirements.

The other is segmented control. Although the speed segmentated control is an open-loop control, when the target speed update frequently, it always has better performance than PID control in our test. As long as the speed corresponding to each PWM value is measured, the corresponding PWM value output can be found instantaneously when the new target speed arrives.

Launchpad is programmed to measure the speed value corresponding to each PWM value automatically. To reduce the movement space, we measured the left and right wheels are alternately, and the speed of the other wheel is set to zero during the measurement.

## V. EXPERIMENT

### A. SLAM

SLAM (Simultaneous Localization and Mapping) is one of the core technologies of mobile robots and has been greatly developed over the years. To test the Plantbot, we mounted RPLIDAR A1 to run gmapping package based on Gmapping algotithm. GMapping [2] is a highly efficient Rao-Blackwellized particle filer to learn grid maps from laser range data. The gmapping package provides laser-based SLAM, as a ROS node called slam_gmapping. Using slam_gmapping, one can create a 2-D occupancy grid map (like a building floorplan) from laser and pose data collected by a mobile robot. In addition to the LiDAR data, slam_gmapping also subscribes to the odometer data, which is calculated by the formula (13), Plantbot provides *odom_node* to calculate the odometer and publish it.

In our experiment, Plantbot can realize SLAM using slam_gmapping package.

### B. Navigation

The other experiment is Navigation. In this experiment, Navigation stack [13] is used to realize the task. Navigation stack is well encapsulated and ROS provides a convenient interface for Navigation. The experimental process and result are shown in Fig. 8. The map was drawn in the previous experiment.

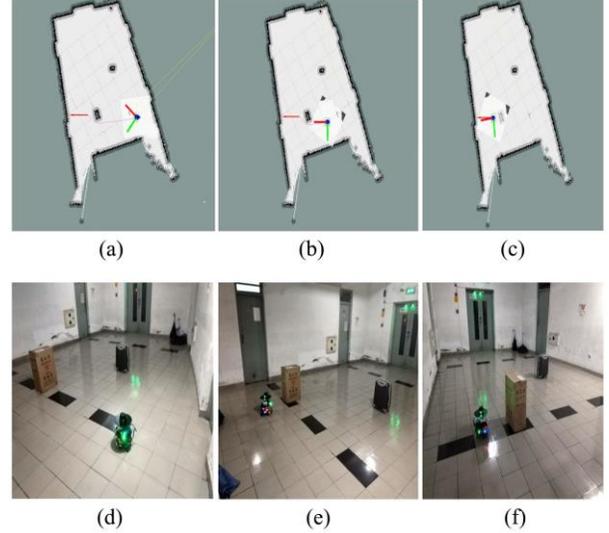

Fig. 8 Navigation

As shown in Fig. 8 (a) (d), the map of the environment and the position of the robot itself in the map were obtained in the SLAM experiment. This information was displayed in rviz, the 3D visualization tool in ROS. The target pose of the robot could also be set in rviz. After the setting was complete, the Navigation stack would automatically plan a reasonable path to guide the main movement of the robot. If the planning path were found to be invalid during execution, such as detecting a moving obstacle, the planner would plan a new path. Fig. 8 (b) (e) shows one of the states of the robot during Navigation. After the robot reached the target with the default accuracy, the robot stopped its movement as shown in Fig. 8(c) (f), and the navigation task finished.

## VI. CONCLUSION

We have built a new robot platform based on ROS, which is easy to reproduce and develop. Plantbot has only 60% of the volume, half of the weight, and one sixth of the price compared to Turtlebot2; and more importantly, it is easy to develop, such as installing new sensors, replacing batteries and controllers, changing the height between support boards, etc. This article analyzes Plantbot in several aspects, including hardware design, kinematics analysis, and low-level control. Low-level control is encapsulated as ROS interface for the easy call. Finally, the SLAM and Navigation experiments were tested on this platform and the ideal results were obtained. The project is open source, convenient for researchers to use and contribute.